\documentclass[11pt,letterpaper]{article}
\usepackage{naaclhlt2018}
\usepackage{times}
\usepackage{latexsym}

\usepackage[utf8]{inputenc} 
\usepackage{booktabs}       
\usepackage{amsfonts}       
\usepackage{nicefrac}       
\usepackage{amsmath}
\usepackage{graphicx}
\usepackage{fancyvrb}
\usepackage{placeins}
\usepackage{listings}
\aclfinalcopy 
\def\aclpaperid{1750} 

\title{Demand-Weighted Completeness Prediction for a Knowledge Base}
\author{Andrew Hopkinson \and Amit Gurdasani \and Dave Palfrey \and Arpit Mittal \\
  Amazon Research Cambridge \\
  Cambridge, UK \\
  {\tt \{hopkia, amitgurd, dpalfrey, mitarpit\}@amazon.co.uk} \\}
\date{13th April 2018}
\begin{document}
\maketitle

\begin{abstract}
In this paper we introduce the notion of Demand-Weighted Completeness, allowing estimation of the completeness of a knowledge base with respect to how it is used. Defining an entity by its classes, we employ usage data to predict the distribution over relations for that entity. For example, instances of \textit{person} in a knowledge base may require a birth date, name and nationality to be considered complete. These predicted relation distributions enable detection of important gaps in the knowledge base, and define the required facts for unseen entities. Such characterisation of the knowledge base can also quantify how usage and completeness change over time. We demonstrate a method to measure Demand-Weighted Completeness, and show that a simple neural network model performs well at this prediction task.
\end{abstract}

\section{Introduction}

Knowledge Bases (KBs) are widely used for representing information in a structured format. Such KBs, including Wikidata~\cite{vrandecic2014}, Google Knowledge Vault~\cite{dong2014}, and YAGO~\cite{Suchanek2007}, often store information as facts in the form of triples, consisting of two entities and a relation between them. KBs have many applications in fields such as machine translation, information retrieval and question answering~\cite{Ferrucci2012}.

When considering a KB's suitability for a task, primary considerations are the number of facts it contains~\cite{farber2015}, and the precision of those facts. One metric which is often overlooked is \textit{completeness}. This can be defined as the proportion of facts about an entity that are present in the KB as compared to an ideal KB which has every fact that can be known about that entity. For example, previous research~\cite{suchanek2011,min2013distant} has shown that between 69\% and 99\% of entities in popular KBs lack at least one relation that other entities in the same class have. As of 2016, Wikidata knows the father of only 2\% of all people in the KB~\cite{Galarraga2017}. Google found that 71\% of people in Freebase have no known place of birth, and 75\% have no known nationality~\cite{dong2014}.

Previous work has focused on a general concept of completeness, where all KB entities are expected to be fully complete, independent of how the KB is used~\cite{Motro1989,razniewski2016, Zaveri2013QualityAF}. This is a problem because different use cases of a KB may have different completeness requirements. For this work, we were interested in determining a KB's completeness with respect to its query usage, which we term \textit{Demand-Weighted Completeness}. For example, a relation used 100 times per day is more important than one only used twice per day.

\subsection{Problem specification}

We define our task as follows:

`Given an entity $E$ in a KB, and query usage data of the KB, predict the distribution of relations that $E$ must have in order for 95\% of queries about $E$ to be answered successfully.'

\subsection{Motivation}

Demand-Weighted Completeness allows us to predict both important missing relations for existing entities, and relations required for unseen entities. As a result we can target acquisition of sources to fill important KB gaps.

It is possible to be entirely reactive when addressing gaps in KB data. Failing queries can be examined and  missing fields marked for investigation. However, this approach assumes that:
\begin{enumerate}
\item the same KB entity will be accessed again in future, making the data acquisition useful. This is far from guaranteed.
\item the KB already contains all entities needed. While this may hold for some use cases, the most useful KB's today grow and change to reflect a changing world.
\end{enumerate}

Both assumptions become unnecessary with an abstract representation of entities, allowing generalization to predict usage. The appropriateness of the abstract representation can be measured by how well the model distinguishes different entity types, and how well the model predicts actual usage for a set of entities, either known or unknown.

Further, the Demand-Weighted Completeness of a KB with respect to a specific task can be used as a metric for system performance at that task. By identifying gaps in the KB, it allows targeting of specific improvements to achieve the greatest increase in completeness.

Our work is the first to consider KB completeness using the distribution of observed KB queries as a signal. This paper details a learning-based approach that predicts the required relation distributions for both seen and unseen class signatures (Section~\ref{sec:representation}), and shows that a neural network model can generalize relation distributions efficiently and accurately compared to a baseline frequency-based approach (Section~\ref{sec:results}). 

\section{Related work}

Previous work has studied the completeness of the individual properties or database tables over which queries are executed~\cite{razniewski2011,Razniewski2015}. This approach is suitable for KBs or use cases where individual tables, and individual rows in those tables, are all of equal importance to the KB, or are queried separately.

Completeness of KBs has also been measured based on the cardinality of properties. \citet{Galarraga2017} and \citet{mirza2016} estimated cardinality for several relations with respect to individual entities, yielding targeted completeness information for specific entities. This approach depends on the availability of relevant free text, and uses handcrafted regular expressions to extract the information, which can be noisy and doesn't scale to large numbers of relations.

The potential for metrics around completeness and dynamicity of a KB are explored in \citet{Zaveri2013QualityAF}, focusing on the task-independent idea of completeness, and the temporal currency, volatility and timeliness of the KB contents. While their concept of timeliness has some similarities to demand-weighted completeness in its task-specific 'data currency', we focus more on how the demand varies over time, and how the completeness of the KB varies with respect to that change in demand. 

\section{Representing Entities}
\label{sec:representation}
\subsection{Class Distributions}
\label{subsec:classdist}
The data for a single entity does not generalize on its own. In order to generalize from observed usage information to unseen entities and unseen usage, and smooth out outliers, we need to combine data from similar entities. Such combination requires a shared entity representation, allowing combination of similar entities while preventing their confusion with dissimilar entities.

For this work, an entity may be a member of multiple classes (or types). We aggregate usage across multiple entities by abstracting to their classes. Membership of a class can be considered as a binary attribute for an entity, with the entity's membership of all the classes considered in the analysis forming a \textit{class signature}.

\begin{figure}
\centering
\begin{lstlisting}
barackObama:
    person:     1
    politician: 1
    democrat:   1
    republican: 0
    writer:     1
\end{lstlisting}
\caption{Class signature for \textit{barackObama}. Other entities with the same class membership will have the same signature.}
\label{ObamaSignature}
\end{figure}

For example, the entity \textit{barackObama} is a \textit{person}, \textit{politician}, \textit{democrat}, and \textit{writer}, among other classes. He is not a \textit{republican}. Considering these five classes as our class space, the class signature for \textit{barackObama} would look like Figure \ref{ObamaSignature}.

Defining an entity by its classes has precedent in previous work~\cite{Galarraga2017,razniewski2016}. It allows consideration of entities and class combinations not yet seen in the KB (though not entirely new classes).

\subsection{Relation Distributions}
\label{subsec:reldist}
\begin{figure}
\centering
\includegraphics[width=\linewidth]{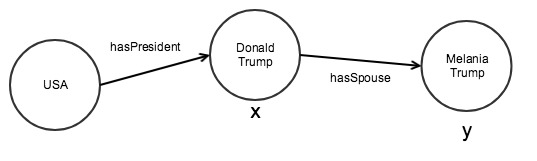}
\caption{Graph representation of the facts needed to solve the query in Equation \ref{EqQuery}. The path walked by the query can branch arbitrarily, but maintains a directionality from initial entities to result entities.}
\label{Query}
\end{figure}

KB queries can be considered as graph traversals, stepping through multiple edges of the knowledge graph to determine the result of multi-clause query. For example, the query:
\begin{equation}
\centering
y: \textrm{hasPresident}(\textrm{USA}, x) \wedge \textrm{hasSpouse}(y, x)
\label{EqQuery}
\end{equation}
determines the spouse of the president of the United States by composing two clauses, as shown in Figure \ref{Query}.

The demand-weighted importance of a relation $R$ for an entity $E$ is defined as the number of query clauses about $E$ which contain $R$, as a fraction of the total number of clauses about $E$. For example, Equation \ref{EqQuery} contains two clauses. As the first clause queries for the hasPresident relation of the USA entity, we attribute this occurrence of hasPresident to the USA entity. Aggregating the clauses for an entity gives a total entity usage of the form seen in Figure \ref{USAUsage}.

\begin{figure}
\centering
\begin{lstlisting}
USA:
    hasPresident:     13
    hasCapital:       8
    hasPopulation:    6
    ...
\end{lstlisting}
\caption{Absolute usage data for the entity \textit{USA}.}
\label{USAUsage}
\end{figure}

Since the distribution of relation usage is dominated by a few high-value relations (see Figure \ref{Overlap}), we only consider relations required to satisfy 95\% of queries.

\subsection{Predicting Relations from Classes}
\label{subsec:relfromcla}

\begin{figure}
\centering
\begin{lstlisting}
barackObama:
    hasHeight:     0.16
    hasBirthdate:  0.12
    hasBirthplace: 0.08
    hasSpouse:     0.07
    hasChild:      0.05
\end{lstlisting}
\caption{An example of a predicted relation distribution for an individual entity. The values represent the proportion of usage of the entity that requires the given relation.}
\label{obamaRels}
\end{figure}

Combining the two representation methods above, we aim to predict the relation distribution for a given entity (as in Figure \ref{obamaRels}) using the class membership for the entity (as in Figure \ref{ObamaSignature}). This provides the expected usage profile of an entity, potentially before it has seen any usage.

\section{Data and Models}
\label{sec:data}
\subsection{Our knowledge base}
\label{subsec:ourkb}

We make use of a proprietary KB \cite{TunstallPedoe10aimag} constructed over several years, combining a hand-curated ontology with publicly available data from Wikipedia, Freebase, DBPedia, and other sources. However, the task can be applied to any KB with usage data, relations and classes. We use a subset of our KB for this analysis due to the limitation of model size as a function of the number of classes (input features) and the number of relations (output features).

Our usage data is generated by our Natural Language Understanding system, which produces KB queries from text utterances. Though it is difficult to remove all biases and errors from the system when operated at industrial scale, we use a hybrid system of curated rules and statistical methods to reduce such problems to a minimum. Such errors should not impact the way we evaluate different models for their ability to model the data itself.

\subsection{Datasets}
\label{subsec:datasets}

To create a class signature, we first determine the binary class membership vector for every entity in the usage dataset. We then group entities by class signature, so entities with identical class membership are grouped together.

For each class signature, we generate the relation distribution from the usage data of the entities with that signature. In our case, this usage data is a random subset of query traffic against the KB taken from a specific period of time. The more usage a class signature has, the more fine-grained the distribution of relations becomes. The data is divided into 10 cross-validation folds to ensure that no class signature appears in both the validation and training sets.

We generate 3 different sizes of dataset for experimentation (see Table \ref{DataTable}), to see how dataset size influences the models.

\begin{table}
\centering
\begin{tabular}{lrrr}
\toprule
Dataset & Classes & Relations & Signatures \\
\midrule
$D1_{small}$ & 4400 & 1300 & 12000 \\
$D2_{medium}$ & 8000 & 2000 & 25000 \\
$D3_{large}$ & 9400 & 2100 & 37000 \\
\bottomrule
\end{tabular}
\caption{Dataset statistics.}
\label{DataTable}
\end{table}

\subsection{Relation prediction models}
\label{sec:models}

\subsubsection{Baseline - Frequency-Based}
In this approach, we compute the relation distribution for each individual class by summing the usage data for all entities of that class (see Section \ref{sec:representation}). This gives a combined raw relation usage as seen in Figure \ref{PersonUsage}.

\begin{figure}
\centering
\begin{lstlisting}
person:
    hasName:     31
    hasAge:      18
    hasHeight:   11
    ...
\end{lstlisting}
\caption{Aggregated usage data for the class \textit{person}.}
\label{PersonUsage}
\end{figure}

For every class in the training set we store this raw relation distribution. At test time, we compute the predicted relation distribution for a class signature as the normalized sum of the raw distributions of all its classes. However, these single-class distributions do not capture the influence of class co-occurrence, where the presence of two classes together may have a stronger influence on the importance of a relation than each class on their own. Additionally, storing distributions for each class signature does not scale, and does not generalize to unseen class combinations.

\subsubsection{Learning-Based Approaches}

To investigate the impact of class co-occurrence, we use two different learning models to predict the relation distribution for a given set of input classes. The vector of classes comprising the class signature is used as input to the learned models.

\textbf{Linear regression.} Using the normalized relation distribution for each class signature, we trained a least-squares linear regression model to predict the relation distribution from a binary vector of classes. This model has $(n \times m)$ parameters, where $n$ is the number of input classes and $m$ is the number of relations. We implemented our linear regression model using Scikit-learn toolkit~\cite{scikit-learn}.

\textbf{Neural network.} We trained a feed-forward neural network using the binary class vector as the input layer, with a low-dimensional ($h$) hidden layer (with rectified linear unit as activation) followed by a softmax output layer of the size of the relation set. This model has $h(n + m)$ parameters, which depending on the value of $h$ is significantly smaller than the linear regression model. The objective function used for training was Kullback-Liebler Divergence. We chose Keras~\cite{chollet2015keras} to implement the neural network model. The model had a single 10-node Rectified Linear Unit hidden layer, with a softmax over the output.

\section{Evaluation}
\begin{figure}
\centering
\includegraphics[width=\linewidth]{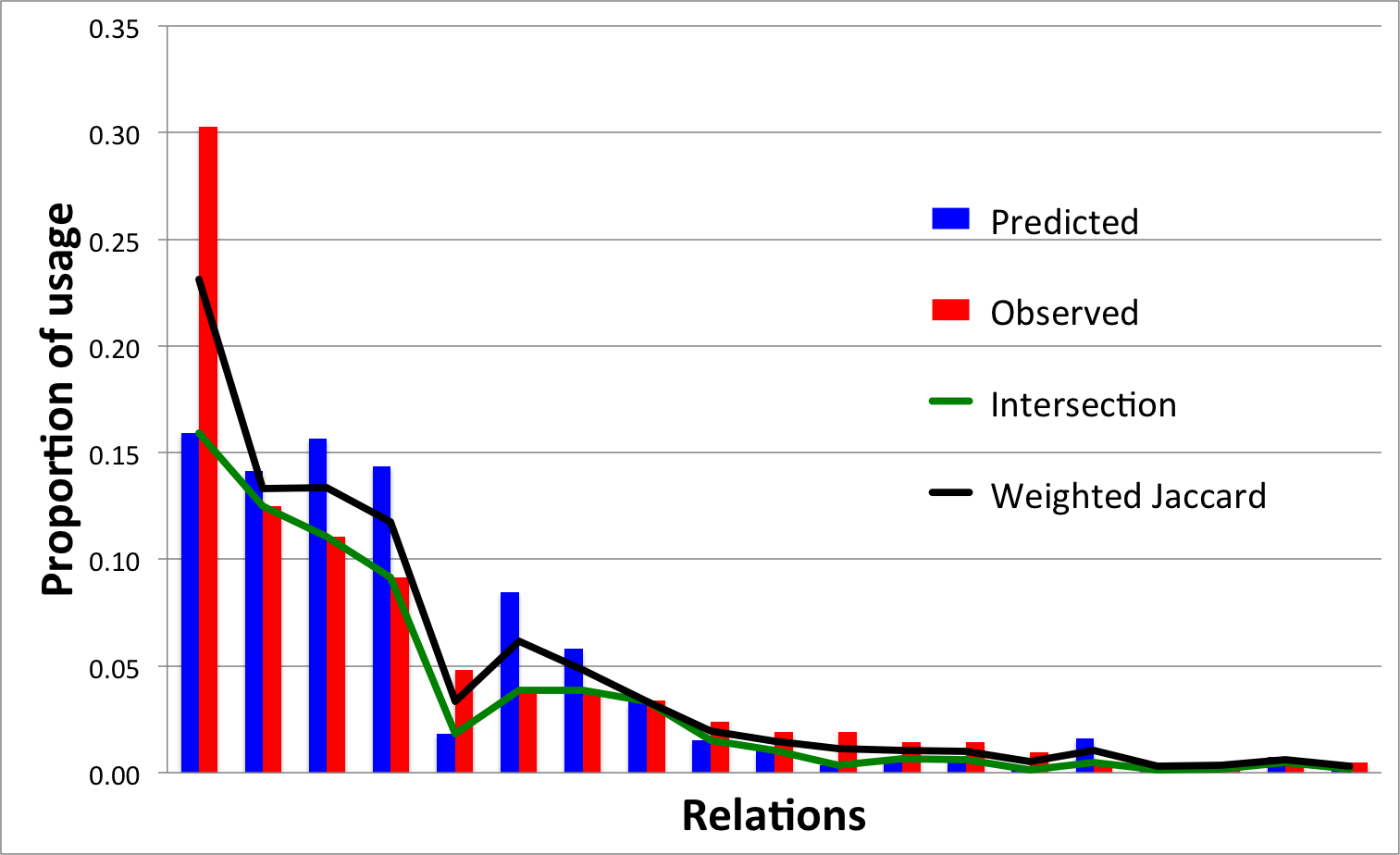}
\caption{Example histogram of the predicted (using a neural model) and observed relation distributions for a single \textit{class signature}, showing the region of intersection in green and the weighted Jaccard index in black.}
\label{Overlap}
\end{figure}

We compare the predicted relation distributions to those observed for the test examples in two ways:

\textbf{Weighted Jaccard Index.} We modified the Jaccard index~\cite{jaccard1912} to include a weighting term, which weights every relation with the mean weight in the predicted and observed distribution (see Figure~\ref{Overlap}). This rewards a correctly predicted relation without focusing on the proportion predicted for that relation, and is sufficient to define a set of important relations for a \textit{class signature}. This is given by:
\begin{equation}
\centering
J = \frac{\sum\limits_{i} W(R_i) \times R_i \in (P \cap O)}{\sum\limits_{i} W(R_i) \times R_i \in (P \cup O)}
\label{Jaccard}
\end{equation}
where $P$ is the predicted distribution, $O$ is the observed distribution, $W(R_i)$ is the mean weight of relation $R_i$ in $P$ and $O$.
We also calculate false negatives (observed but not predicted) and false positives (predicted but not observed), by modifying the second term in the numerator of Equation \ref{Jaccard} to give $P \backslash O$ and $O \backslash P$, rather than $P \cap O$.

\textbf{Intersection.} We compute the intersection of the two distributions (see Figure \ref{Overlap}). This is a more strict comparison between the distributions which penalizes differences in weight for individual relations. This is given by:
\begin{equation}
I = \sum_i min(P(R_i), O(R_i))
\end{equation}

\subsection{Usage Weighted Evaluation}
\label{subsec:usageWeighting}

We also evaluated the models using the Weighted Jaccard index and Intersection methods, but weighting by usage counts for each signature. This metric rewards the models more for correctly predicting relation distributions for common class signatures in the usage data. While unweighted analysis is useful to examine how the model covers the breadth of the problem space, weighted evaluation more closely reflects the model's utility for real usage data.

\subsection{Temporal Prediction}
\label{temporalPrediction}

Additionally, we evaluated the models on their ability to predict future usage. With an unchanging usage pattern, evaluation against future usage would be equivalent to cross-validation (assuming the same signature distribution in the folds). However, in many real world cases, usage of a KB varies over time, seasonally or as a result of changing user requirements.

Therefore we also evaluated a neural model against future usage data to measure how elapsed time affected model performance. The datasets T1, T2, and T3 each contain 3 datasets (of similar size to $\textrm{D1}_{small}$, $\textrm{D2}_{medium}$, and $\textrm{D1}_{large}$), and were created using usage data from time periods with a fixed offset, $t$.  The base set was created at time $t_0$, T1 at time $t_0 + t$, T2 at time $t_0 + 2t$, and T3 at time $t_0 + 3t$. A time interval was chosen that reflected the known variability of the usage data, such that we would expect the usage to not be the same. 

\section{Results}
\label{sec:results}
\subsection{Cross-Validation}
\label{subsec:cv}
10-fold cross-validation results are shown in Table \ref{ResultsComparison}. The neural network model performs best, outperforming the baseline model by 6-8 percentage points. The regression model performs worst, trailing the baseline model by 4-8 percentage points.

\begin{table}
\centering
\begin{tabular}{lccc}
\toprule
Model & Jaccard & False Neg. & False Pos. \\
\midrule
$D1_{small}$ & & & \\
\midrule
Freq. & 0.604 & 0.084 & 0.311 \\
Regr. & 0.522 & 0.102 & 0.376 \\
NN & \textbf{0.661} & 0.036 & 0.303 \\
\midrule
$D2_{medium}$ & & & \\
\midrule
Freq. & 0.611 & 0.101 & 0.287 \\
Regr. & 0.557 & 0.084 & 0.358 \\
NN & \textbf{0.687} & 0.035 & 0.278 \\
\midrule
$D3_{large}$ & & & \\
\midrule
Freq. & 0.616 & 0.105 & 0.278 \\
Regr. & 0.573 & 0.080 & 0.347 \\
NN & \textbf{0.700} & 0.034 & 0.266 \\
\bottomrule
\end{tabular}
\caption{Unweighted results for the three models on the three datasets.}
\label{ResultsComparison}
\end{table}

\subsubsection{Baseline}
\label{subsubsec:FreqResults}
The baseline model shows little improvement with increasing amounts of data - the results from $\textrm{D1}_{small}$ to $\textrm{D3}_{large}$ (3x more data points) only improve by just over 1 percentage point. This suggests that this model is unable to generalise from the data, which is expected from the lack of class co-occurrence information in the model. Interestingly, the baseline model shows an increase in false negatives on the larger datasets, implying the lack of generalisation is more problematic for more fine-grained relation distributions.

\subsubsection{Linear Regression}
\label{subsubsec:RegressionResults}
The linear regression model gives a much lower Jaccard measure than the baseline model. This is likely due to the number of parameters in the model relative to the number of examples. For $\textrm{D1}_{small}$, the model has approximately 6m parameters, with 12k training examples, making this an under-determined system. For $\textrm{D3}_{large}$ the number of parameters rises to 20m, with 37k training examples, maintaining the poor example:parameter ratio. From this we might expect the performance of the model to be invariant with the amount of data.

However, the larger datasets also have higher resolution relation distributions, as they are aggregated from more individual examples. This has the effect of reducing the impact of outliers in the data, giving improved predictions when the model generalises. We do indeed see that the linear regression model improves notably with larger datasets, closing the gap to the baseline model from 8 percentage points to 4.

\subsubsection{Neural Network}
\label{subsubsec:NeuralResults}
The neural network model shows much better performance than either of the other two methods. The Jaccard score is consistently 6-8\% above the regression model, with far fewer false negatives and smaller numbers of false positives. This is likely to be due to the smaller number of parameters of the neural model versus the linear regression model. For $\textrm{D3}_{large}$, the 10-node hidden layer model amounts to 115k parameters with 37k training examples, a far better ratio (though still not ideal) than for the linear regression model.

\subsubsection{Weighted Evaluation}

We include in Table \ref{WeightedResultsComparison} the results using the weighted evaluation scheme described in Section \ref{subsec:usageWeighting}. This gives more usage-focused evaluation, emphasizing the non-uniform usage of different class signatures. The $\textrm{D3}_{large}$ neural model achieves 85\% precision with a weighted evaluation. With the low rate of false negatives, this indicates that a similar model could be used to predict the necessary relations for KB usage.

\begin{table}
\centering
\begin{tabular}{lccc}
\toprule
Model & Jaccard & False Neg. & False Pos. \\
\midrule
$D1_{small}$ & & & \\
\midrule
Freq. & 0.779 & 0.066 & 0.123 \\
Regr. & 0.667 & 0.090 & 0.242 \\
NN & \textbf{0.808} & 0.032 & 0.159 \\
\midrule
$D2_{medium}$ & & & \\
\midrule
Freq. & 0.816 & 0.059 & 0.094 \\
Regr. & 0.703 & 0.077 & 0.220 \\
NN & \textbf{0.840} & 0.037 & 0.123 \\
\midrule
$D3_{large}$ & & & \\
\midrule
Freq. & 0.819 & 0.062 & 0.088 \\
Regr. & 0.720 & 0.069 & 0.210 \\
NN & \textbf{0.850} & 0.038 & 0.113 \\
\bottomrule
\end{tabular}
\caption{Usage-weighted results for the three models on the three datasets.}
\label{WeightedResultsComparison}
\end{table}

\subsection{Intersection}
\label{subsec:Intersection}

Table \ref{OverlapTable} gives measurements of the intersection metric. These show a similar trend to the Jaccard scores, with lower absolute values from the stricter evaluation metric. Although the Jaccard measure shows correct relation set prediction with a precision of 0.700, predicting the proportions for those relations accurately remains a difficult problem. The best value we achieved was 0.398.

\begin{table}
\centering
\begin{tabular}{lccc}
\toprule
Model & Freq. & Regr. & NN \\
\midrule
Inter. & 0.319 & 0.278 & 0.398 \\
\bottomrule
\end{tabular}
\caption{Results for the three methods for the $\textrm{D3}_{large}$ dataset using the intersection metric. The difference between the methods is similar to the Jaccard measure above.}
\label{OverlapTable}
\end{table}

\subsection{Unweighted Temporal Prediction}
\label{subsec:temporal}

\begin{table}
\centering
\begin{tabular}{lccc}
\toprule
Interval & T1 & T2 & T3 \\
\midrule
$D1_{small}$ & 0.661 & 0.659 & 0.657 \\
$D2_{medium}$ & 0.705 & 0.699 & 0.696 \\
$D3_{large}$ & 0.712 & 0.708 & 0.704 \\
\bottomrule
\end{tabular}
\caption{Results of training a neural model on all available data for $\textrm{D1}_{small}$ - $\textrm{D3}_{large}$, then evaluating on T1-T3. The values for $\textrm{D2}_{medium}$ and $\textrm{D3}_{large}$ are higher than cross-validation, as cross-validation never tests a model on examples used to train it. However, the T datasets contain all data from the specified period. The downward trend with increasing T is clear, but slight.}
\label{TemporalResults}
\end{table}

In addition to evaluating models on their ability to predict the behaviour of unseen class signatures, we also evaluated the neural model on its ability to predict future usage behaviour. The results of this experiment are given in Table \ref{TemporalResults}.

We observe a very slight downward trend in the precision of the model using all three base datasets ($\textrm{D1}_{small}$ - $\textrm{D3}_{large}$), with a steeper (but still slight) downward trend for the larger datasets. This suggests that a model trained on usage data from one period of time will have significant predictive power on future datasets. 

\section{Measuring Completeness of a KB}

Once we have a suitable model of the expected relation distributions for class combinations, we use the model to predict the expected relation distribution for specific entities in our KB. We then compare the predicted relation distribution to the observed relations for each specific entity. The completeness of an entity is given by the sum of the relation proportions for the predicted relations the entity has in the KB.

Any gaps for an entity represent relations that, if added to the KB, would have a quantifiable positive impact on the performance of the KB. By focussing on the most important entities according to our usage, we can target fact addition to have the greatest impact to the usage the KB receives.

By aggregating the completeness values for a set of entities, we may estimate the completeness of subsets of the KB. This aggregation is weighted by the frequency with which the entity appears in the usage data, giving a usage-weighted measure of the subset's completeness. These subsets can represent individual topics, individual classes of entity, or overall information about the KB as a whole.

For example, using the best neural model above on an unrepresentative subset of our KB, we evaluate the completeness of that subset at 58.3\%. This not only implies that we are missing a substantial amount of necessary information for these entities with respect to the usage data chosen, but permits targeting of source acquisition to improve the entity completness in aggregate. For example, if we are missing a large number of \textit{hasBirthdate} facts for people, we might locate a source that has that information. We can quantify the benefit of that effort in terms of improved usage performance.

\section{Conclusions and Future Work}

We have introduced the notion of Demand-Weighted Completeness as a way of determining a KB's suitability by employing usage data. We have demonstrated a method to predict the distribution of relations needed in a KB for entities of a given class signature, and have compared three different models for predicting these distributions. Further, we have described a method to measure the completeness of a KB using these distributions.

For future work we would like to try complex neural network architectures, regularisation, and semantic embeddings or other abstracted relations to enhance the signatures. We would also like to investigate Good-Turing frequency estimation~\cite{good1953}.

\bibliography{naacl-industry-2018}

\begin{thebibliography}{}
\expandafter\ifx\csname natexlab\endcsname\relax\def\natexlab#1{#1}\fi

\bibitem[{Chollet(2015)}]{chollet2015keras}
Fran\c{c}ois Chollet. 2015.
\newblock Keras.
\newblock \url{https://github.com/fchollet/keras}.

\bibitem[{Dong et~al.(2014)Dong, Gabrilovich, Heitz, Horn, Lao, Murphy,
  Strohmann, Sun, and Zhang}]{dong2014}
Xin Dong, Evgeniy Gabrilovich, Geremy Heitz, Wilko Horn, Ni~Lao, Kevin Murphy,
  Thomas Strohmann, Shaohua Sun, and Wei Zhang. 2014.
\newblock \href{https://doi.org/10.1145/2623330.2623623}{Knowledge {V}ault: A
  web-scale approach to probabilistic knowledge fusion}.
\newblock In {\em Proceedings of the 20th ACM SIGKDD International Conference
  on Knowledge Discovery and Data Mining\/}. ACM, New York, NY, USA, KDD '14,
  pages 601--610.
\newblock \url{https://doi.org/10.1145/2623330.2623623}.

\bibitem[{F{\"a}rber et~al.(2015)F{\"a}rber, Ell, Menne, and
  Rettinger}]{farber2015}
Michael F{\"a}rber, Basil Ell, Carsten Menne, and Achim Rettinger. 2015.
\newblock A comparative survey of {DB}pedia, {F}reebase, {O}pen{C}yc,
  {W}ikidata, and {YAGO}.
\newblock {\em Semantic Web Journal, July\/} .

\bibitem[{Ferrucci(2012)}]{Ferrucci2012}
D.~A. Ferrucci. 2012.
\newblock \href{https://doi.org/10.1147/JRD.2012.2184356}{Introduction to "this
  is watson"}.
\newblock {\em IBM J. Res. Dev.\/} 56(3):235--249.
\newblock \url{https://doi.org/10.1147/JRD.2012.2184356}.

\bibitem[{Gal\'{a}rraga et~al.(2017)Gal\'{a}rraga, Razniewski, Amarilli, and
  Suchanek}]{Galarraga2017}
Luis Gal\'{a}rraga, Simon Razniewski, Antoine Amarilli, and Fabian~M. Suchanek.
  2017.
\newblock \href{https://doi.org/10.1145/3018661.3018739}{Predicting
  completeness in knowledge bases}.
\newblock In {\em Proceedings of the Tenth ACM International Conference on Web
  Search and Data Mining\/}. ACM, New York, NY, USA, WSDM '17, pages 375--383.
\newblock \url{https://doi.org/10.1145/3018661.3018739}.

\bibitem[{Good(1953)}]{good1953}
I.~J. Good. 1953.
\newblock \href{https://doi.org/10.1093/biomet/40.3-4.237}{The population
  frequencies of species and the estimation of population parameters}.
\newblock {\em Biometrika\/} 40(3-4):237.
\newblock \url{https://doi.org/10.1093/biomet/40.3-4.237}.

\bibitem[{Jaccard(1912)}]{jaccard1912}
Paul Jaccard. 1912.
\newblock \href{https://doi.org/10.1111/j.1469-8137.1912.tb05611.x}{The
  distribution of the flora in the alpine zone.1}.
\newblock {\em New Phytologist\/} 11(2):37--50.
\newblock \url{https://doi.org/10.1111/j.1469-8137.1912.tb05611.x}.

\bibitem[{Min et~al.(2013)Min, Grishman, Wan, Wang, and
  Gondek}]{min2013distant}
Bonan Min, Ralph Grishman, Li~Wan, Chang Wang, and David Gondek. 2013.
\newblock Distant supervision for relation extraction with an incomplete
  knowledge base.
\newblock In {\em HLT-NAACL\/}. pages 777--782.

\bibitem[{Mirza et~al.(2016)Mirza, Razniewski, and Nutt}]{mirza2016}
Paramita Mirza, Simon Razniewski, and Werner Nutt. 2016.
\newblock Expanding {W}ikidata's parenthood information by 178\%, or how to
  mine relation cardinalities.
\newblock In {\em ISWC 2016 Posters \& Demonstrations Trac\/}. CEUR-WS. org.

\bibitem[{Motro(1989)}]{Motro1989}
Amihai Motro. 1989.
\newblock \href{https://doi.org/10.1145/76902.76904}{Integrity = validity +
  completeness}.
\newblock {\em ACM Trans. Database Syst.\/} 14(4):480--502.
\newblock \url{https://doi.org/10.1145/76902.76904}.

\bibitem[{Pedregosa et~al.(2011)Pedregosa, Varoquaux, Gramfort, Michel,
  Thirion, Grisel, Blondel, Prettenhofer, Weiss, Dubourg, Vanderplas, Passos,
  Cournapeau, Brucher, Perrot, and Duchesnay}]{scikit-learn}
F.~Pedregosa, G.~Varoquaux, A.~Gramfort, V.~Michel, B.~Thirion, O.~Grisel,
  M.~Blondel, P.~Prettenhofer, R.~Weiss, V.~Dubourg, J.~Vanderplas, A.~Passos,
  D.~Cournapeau, M.~Brucher, M.~Perrot, and E.~Duchesnay. 2011.
\newblock Scikit-learn: Machine learning in {P}ython.
\newblock {\em Journal of Machine Learning Research\/} 12:2825--2830.

\bibitem[{Razniewski et~al.(2015)Razniewski, Korn, Nutt, and
  Srivastava}]{Razniewski2015}
Simon Razniewski, Flip Korn, Werner Nutt, and Divesh Srivastava. 2015.
\newblock \href{https://doi.org/10.1145/2723372.2750544}{Identifying the extent
  of completeness of query answers over partially complete databases}.
\newblock In {\em Proceedings of the 2015 ACM SIGMOD International Conference
  on Management of Data\/}. ACM, New York, NY, USA, SIGMOD '15, pages 561--576.
\newblock \url{https://doi.org/10.1145/2723372.2750544}.

\bibitem[{Razniewski and Nutt(2011)}]{razniewski2011}
Simon Razniewski and Werner Nutt. 2011.
\newblock Completeness of queries over incomplete databases.
\newblock {\em VLDB\/} 4:749--760.

\bibitem[{Razniewski et~al.(2016)Razniewski, Suchanek, and
  Nutt}]{razniewski2016}
Simon Razniewski, Fabian~M Suchanek, and Werner Nutt. 2016.
\newblock But what do we actually know.
\newblock {\em Proceedings of AKBC\/} pages 40--44.

\bibitem[{Suchanek et~al.(2011)Suchanek, Gross-Amblard, and
  Abiteboul}]{suchanek2011}
Fabian Suchanek, David Gross-Amblard, and Serge Abiteboul. 2011.
\newblock Watermarking for ontologies.
\newblock {\em The semantic web--ISWC 2011\/} pages 697--713.

\bibitem[{Suchanek et~al.(2007)Suchanek, Kasneci, and Weikum}]{Suchanek2007}
Fabian~M. Suchanek, Gjergji Kasneci, and Gerhard Weikum. 2007.
\newblock \href{https://doi.org/10.1145/1242572.1242667}{{YAGO}: {A} core of
  semantic knowledge}.
\newblock In {\em Proceedings of the 16th International Conference on World
  Wide Web\/}. ACM, New York, NY, USA, WWW '07, pages 697--706.
\newblock \url{https://doi.org/10.1145/1242572.1242667}.

\bibitem[{Tunstall-Pedoe(2010)}]{TunstallPedoe10aimag}
William Tunstall-Pedoe. 2010.
\newblock \href{https://doi.org/10.1609/aimag.v31i3.2298}{{True Knowledge}:
  Open-domain question answering using structured knowledge and inference}.
\newblock {\em AI Magazine\/} 31(3):80--92.
\newblock \url{https://doi.org/10.1609/aimag.v31i3.2298}.

\bibitem[{Vrande\v{c}i\'{c} and Kr\"{o}tzsch(2014)}]{vrandecic2014}
Denny Vrande\v{c}i\'{c} and Markus Kr\"{o}tzsch. 2014.
\newblock \href{https://doi.org/10.1145/2629489}{Wikidata: A free collaborative
  knowledgebase}.
\newblock {\em Commun. ACM\/} 57(10):78--85.
\newblock \url{https://doi.org/10.1145/2629489}.

\bibitem[{Zaveri et~al.(2013)Zaveri, Rula, Maurino, Pietrobon, Lehmann, and
  Auer}]{Zaveri2013QualityAF}
Amrapali Zaveri, Anisa Rula, Andrea Maurino, Ricardo Pietrobon, Jens Lehmann,
  and S{\"o}ren Auer. 2013.
\newblock Quality assessment for linked open data: A survey.

\end{thebibliography}
\bibliographystyle{acl_natbib}

\end{document}